\documentclass[conference]{IEEEtran}
\IEEEoverridecommandlockouts
\usepackage{cite}
\usepackage{amsmath,amssymb,amsfonts}
\usepackage{algorithmic}
\usepackage[pdftex]{graphicx}
\usepackage{textcomp}
\usepackage{comment}
\def\BibTeX{{\rm B\kern-.05em{\sc i\kern-.025em b}\kern-.08em
    T\kern-.1667em\lower.7ex\hbox{E}\kern-.125emX}}

\newcommand{\oomoji}[1]{{\mbox{\boldmath $#1$}}}

\newcommand{\diffH}[1]{[#1]_{\rm H}}
\newcommand{\diffV}[1]{[#1]_{\rm V}}

\begin{document}

\title{HOG feature extraction from encrypted images for privacy-preserving machine learning\\
}

\author{\IEEEauthorblockN{1\textsuperscript{st} Masaki Kitayama}
\IEEEauthorblockA{\textit{Tokyo Metropolitan University} \\
Tokyo, Japan \\
kitayama-masaki@ed.tmu.ac.jp}
\and
\IEEEauthorblockN{2\textsuperscript{nd} Hitoshi Kiya}
\IEEEauthorblockA{\textit{Tokyo Metropolitan University} \\
Tokyo, Japan \\
kiya@tmu.ac.jp}
}

\maketitle

\begin{abstract}
In this paper, we propose an extraction method of HOG (histograms-of-oriented-gradients) features from encryption-then-compression (EtC) images for privacy-preserving machine learning, 
where EtC images are images encrypted by a block-based encryption method proposed for EtC systems with JPEG compression, 
and HOG is a feature descriptor used in computer vision for the purpose of object detection and image classification.
Recently, cloud computing and machine learning have been spreading in many fields.
However, the cloud computing has serious privacy issues for end users, due to unreliability of providers and some accidents.
Accordingly, we propose a novel block-based extraction method of HOG features, 
and the proposed method enables us to carry out any machine learning algorithms without any influence,
under some conditions.
In an experiment, the proposed method is applied to a face image recognition problem under the use of two kinds of classifiers: 
linear support vector machine (SVM), gaussian SVM, to demonstrate the effectiveness.
\end{abstract}

\begin{IEEEkeywords}
encryption-then-compression， block-based encryption, histgram of oriented gradients
\end{IEEEkeywords}

\section{Introduction}
\label{sec:intro}

In recent years, cloud computing and edge computing that use resourses of providers are rapidly spreading in many fields.
However, there are some security concerns such as unauthorized use and leak of data, and privacy compromise\cite{cloudsecuritysurvey}.
For solving these privacy concerns for end-users, many researches has been conducted\cite{cloudsecurity1,homomorphic1,HOGhomomorphic1,securitydeeplearning1}.

In this paper, we propose a scheme of histgram-of-oriented-gradients (HOG) feature extraction from encryption-then-compression (EtC) images, 
where HOG features are well-known features used mainly in computer vision\cite{HOG}, 
and EtC images are images encrypted by a block-based encryption method\cite{
EtC1,EtC2,EtC3,EtC4,EtCsecurity1,EtC5,securitydeeplearning6}.
EtC images have been applied to privacy preserving machine learning algorithms, 
but HOG features are not extracted yet from EtC ones\cite{EtC6}.

In the proposed method, we encrypt both training and test images by using the same key, then extract HOG features from the encrypted images and apply these features to machine learning algorithms.
HOG feature extracted by using the proposed method is exact the same as those extracted from unencrypted images.
In other words, the encryption gives no influence to the performance of machine learning algorithms with HOG features, 
while protecting visual information of images.
In an experiment, the proposed method is experimentally demonstraded to be effective for a face image recognition problem with a machine learning algorithm such as SVM algorithms.

\section{EtC Images}
\label{EtC}

\begin{figure}[t]
 \begin{minipage}{0.47\hsize}
  \begin{center}
   \includegraphics[scale=0.28]{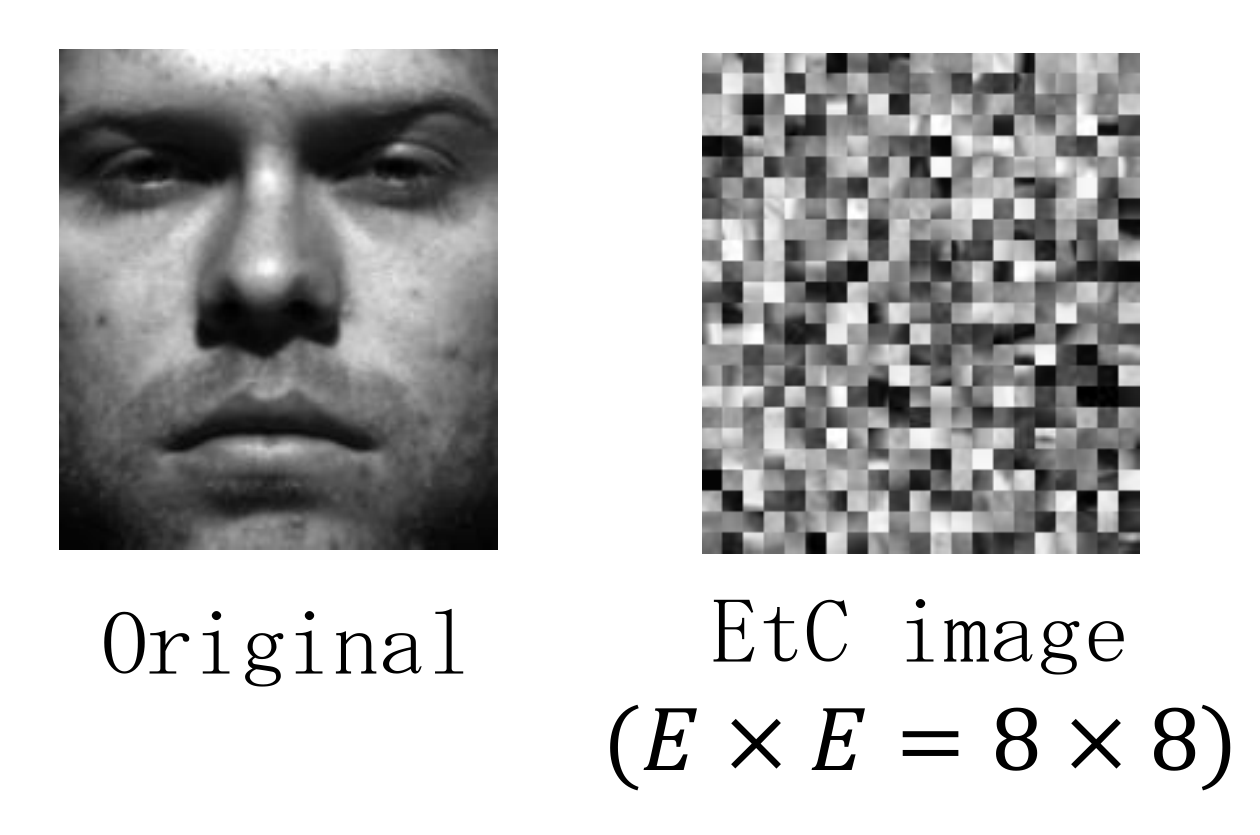}
   \caption{Example of EtC images \label{pictures}}
  \end{center}
 \end{minipage}
 \hfill
 \begin{minipage}{0.47\hsize}
  \begin{center}
   \includegraphics[scale=0.25]{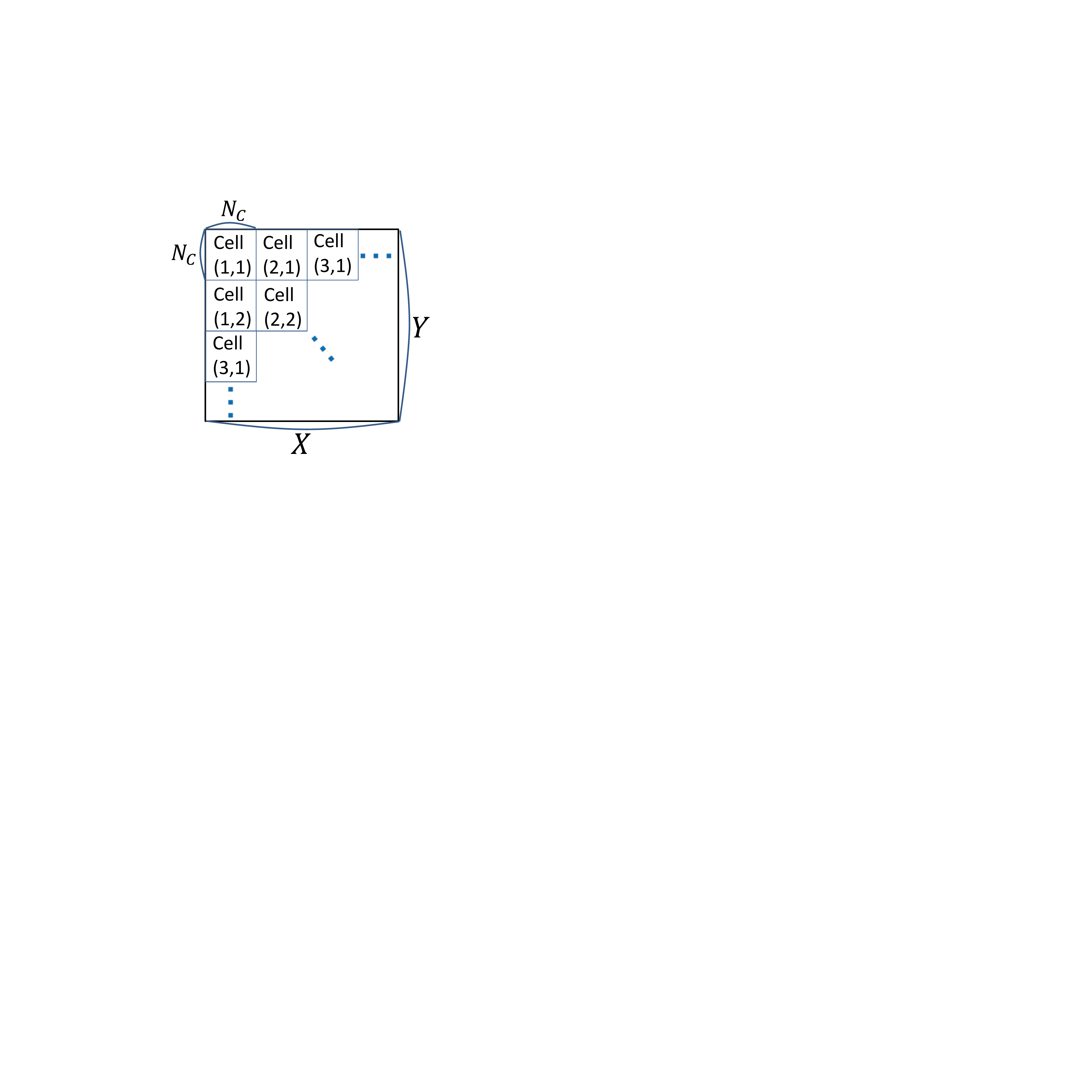}
   \caption{HOG cell separation.  \label{cell}}
  \end{center}
 \end{minipage}
\end{figure}

We focus on EtC images which have been proposed for encryption-then-compression systems with JPEG compression\cite{EtC5,EtCsecurity1}.
EtC images have not only almost the same compression performance as that of unencrypted images, but also enough robustness against various ciphertext-only attacks including jigsaw puzzle solver attacks\cite{EtCsecurity1}.

The procedure of generating EtC images is summarized here.

\noindent 1) Divide an image $\oomoji{L}$ with $X\times Y$ pixels into blocks $\oomoji{B}_m,m=1,2,\ldots,M$ with $E\times E$ pixels, and permute randomly the divided blocks using a random integer generated by a secret key $K_1$, 
where $M$ is the number of blocks in image $\oomoji{L}$.

\noindent 2) Rotate and invert randomly each block using a random integer generated by a key $K_2$.

\noindent 3) Apply the negative-positive transformation to each block using a random binary integer generated by a key $K_3$.
In this step, a transformed pixel value in $i$th block, $p'$ is computed by

\footnotesize
\begin{equation}
\begin{cases}
p'=p    & (r(i)=0) \\
p'=255-p & (r(i)=1)
\end{cases},
\label{negaposi}
\end{equation}
\normalsize
where $r(i)$ is a random binary integer generated by $K_3$ under the probability $P(r(i))=0.5$ and $p$ is the pixel value of an original image with 8 bit per pixels (see Fig.\ref{pictures}).

\section{Proposed method}
\label{sec:hog}

In accordance with the procedure in Sec. \ref{EtC}, image $\oomoji{L}$ is encrypted to generate an EtC image $\oomoji{\hat{L}}$ which consists of $M$ encrypted blocks $\oomoji{\hat{B}}_m$.
Next, HOG features are extracted from the EtC image $\oomoji{\hat{L}}$ as follows.

\begin{figure}[t]
  \begin{center}
   \includegraphics[scale=0.18]{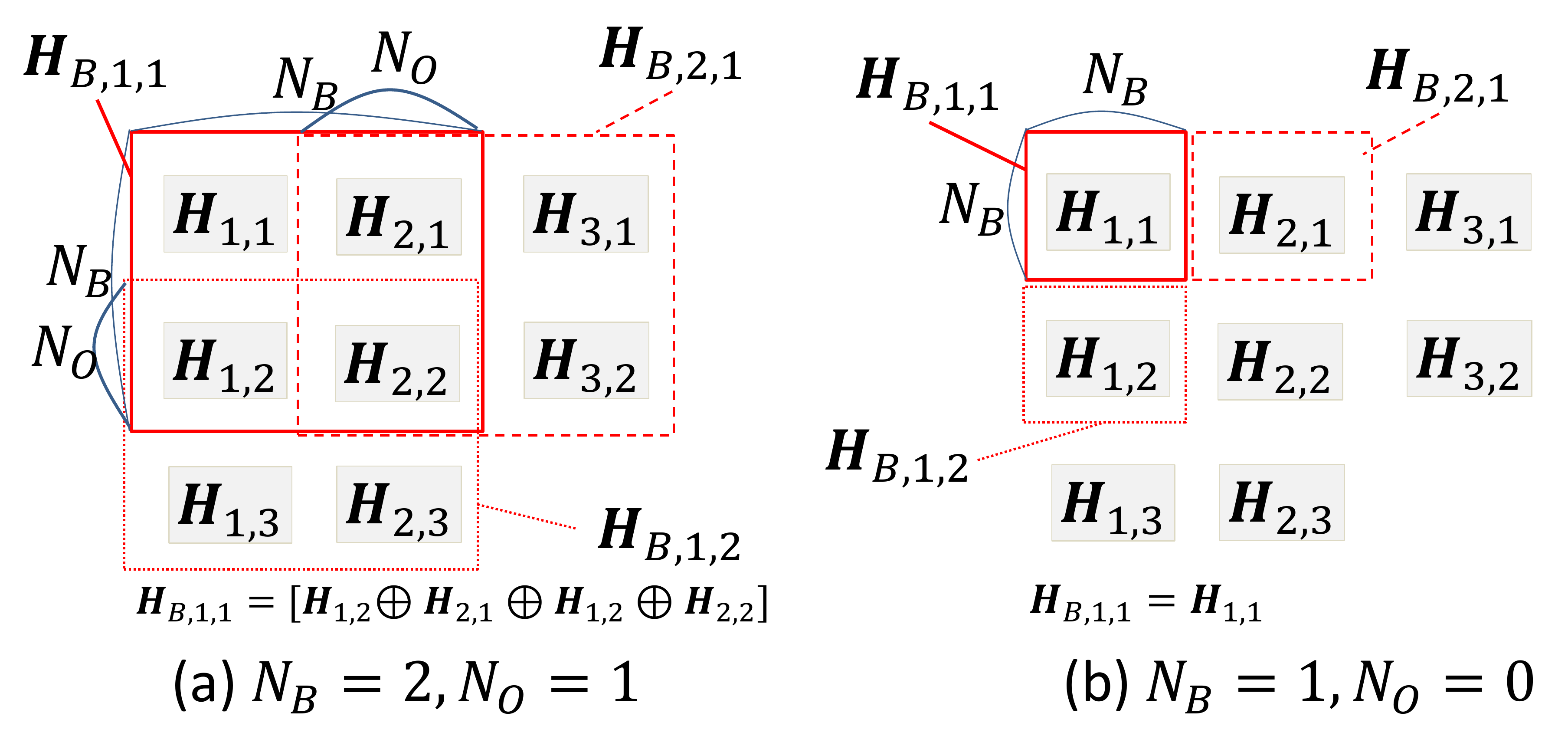}
   \caption{HOG block. $\oplus$ indicates the concatenation of vectors.
\label{cell_block}}
  \end{center}
\end{figure}

\subsubsection*{Step1.\quad Differential Images}

$\oomoji{\hat{L}}$ is devided into $M^\prime$ small grids $\oomoji{\hat{B}}^\prime_m,m=1,2,\ldots,M^\prime$ with $E^\prime\times E^\prime$ pixels.
A horizontal differentiating function $\diffH{\cdot}$ and 
a vertical differentiating function $\diffV{\cdot}$ are respectively defined.
Then, a horizontal differential image $\diffH{\oomoji{\hat{B}}^\prime_{m}}$ and
 a vertical differential image $\diffV{\oomoji{\hat{B}}^\prime_{m}}$ are calculated from each EtC block $\oomoji{\hat{B}}^\prime_m$ respectively as

\footnotesize
\begin{align}
 \label{dif img x}
 \diffH{\hat{B}^\prime_{m}}&(x,y)= \nonumber \\
 & \left\{
 \begin{aligned}
  &\hat{B}^\prime_m(x+1,y)-\hat{B}^\prime_m(x-1,y)&&(2\leq x\leq E^\prime-1)\\
  &\hat{B}^\prime_m(2,y)-\hat{B}^\prime_m(1,y)&&(x=1) \\
  &\hat{B}^\prime_m(E^\prime,y)-\hat{B}^\prime_m(E^\prime-1,y)&&(x=E^\prime)\quad,\\
 \end{aligned}
 \right.
\end{align}
\begin{align}
 \label{dif img y}
 \diffV{\hat{B}^\prime_{m}}&(x,y)= \nonumber \\
 & \left\{
 \begin{aligned}
  &\hat{B}^\prime_m(x,y+1)-\hat{B}^\prime_m(x,y-1)&&(2\leq y\leq E^\prime-1)\\
  &\hat{B}^\prime_m(x,2)-\hat{B}^\prime_m(x,1)&&(y=1)\\
  &\hat{B}^\prime_m(x,E^\prime)-\hat{B}^\prime_m(x,E^\prime-1)&&(y=E^\prime)\quad,\\
 \end{aligned}
 \right.
\end{align}
\normalsize
where $\diffH{\hat{B}^\prime_{m}}(x,y),\diffV{\hat{B}^\prime_{m}}(x,y)$
 and $\hat{B}^\prime_m(x,y)$ 
are pixel values of $\diffH{\oomoji{\hat{B}}^\prime_{m}},\diffV{\oomoji{\hat{B}}^\prime_{m}}$ 
and $\oomoji{\hat{B}}^\prime_m$ respectively 
at a coord $(x,y),x\in\{1,2,\ldots,E^\prime\},y\in\{1,2,\ldots,E^\prime\}$.
%$\{\diffH{\oomoji{\hat{B}}^\prime_{m}}\}$ are combined into 
%one horizontal differential image $\oomoji{l}_H$ with $X\times Y$ pixels.
%$\{\diffV{\oomoji{\hat{B}}^\prime_{m}}\}$ are also combined into $\oomoji{l}_V$.

\subsubsection*{Step2.\quad Gradient Direction and Strength Maps}

A gradient direction map $\oomoji{\theta}_m$ and a gradient strength map $\oomoji{I}_m$ are computed from the $\diffH{\oomoji{\hat{B}}_m}$ and $\diffV{\oomoji{\hat{B}}_m}$, respectively, according to the equations

\footnotesize
\begin{align}
 I_m(x,y)&=\sqrt{\diffH{\hat{B}_m}^{2}(x,y)+\diffV{\hat{B}_m}^{2}(x,y)}\label{I}\\
 \theta_m(x,y)&=\tan^{-1}{\frac{\diffV{\hat{B}_m}(x,y)}{\diffH{\hat{B}_m}(x,y)}},\quad(0<\theta_m(x,y)\leq\pi)\label{theta}\qquad,
\end{align}
\normalsize
where $I_m(x,y)$ and $\theta_m(x,y)$ are pixel values of $\oomoji{I}_m$ and 
$\oomoji{\theta}_m$ respectively at a coord $(x,y)$.
$\{\oomoji{I}_m\}$ are combined into 
one gradient strength map $\oomoji{I}$ with $X\times Y$ pixels.
$\{\oomoji{\theta}_m\}$ are also combined into $\oomoji{\theta}$.

\subsubsection*{Step3.\quad Gradient Histograms}

As shown in Fig.\ref{cell}, maps $\oomoji{I}$ and $\oomoji{\theta}$ are commonly divided into small grids called "HOG cells" with $N_{C}\times N_{C}$ pixels, where the parameter $N_{C}$ is reffered to as "HOG cell size".
First, $\theta(x,y)$ is quantized to angles $\frac{\pi}{2N},\frac{3\pi}{2N},\frac{5\pi}{2N},\ldots,\frac{(2N-1)\pi}{2N}$, where $N$ is the quantization level of $\oomoji{\theta}$.
Then, a gradient histogram $\oomoji{H}_{i,j},i=1,2,\ldots,\frac{X}{N_{C}},j=1,2,\ldots,\frac{Y}{N_{C}}$, which is a histogram of $\theta(x,y)$ in a cell, is made up.
The votes are weighted by $I(x,y)$.

\subsubsection*{Step4.\quad HOG Blocks}

Let us define HOG block $\oomoji{H}_{B,i,j}$ as a group of one or multiple HOG cells (see Fig.\ref{cell_block}).
$\oomoji{H}_{i,j}\oplus \oomoji{H}_{i,j+1}$ in Fig.\ref{cell_block} means the concatenation of $\oomoji{H}_{i,j}$ and $\oomoji{H}_{i,j+1}$.
Each HOG block consists of $N_B\times N_B$ HOG cells, where the parameter $N_B$ is reffered to as ``HOG block size'', and HOG blocks may have the overlap with $N_O$ cells each other.

$\oomoji{H}_{B,i,j}$ has $N\times N_B^{2}$ elements as

\footnotesize
\begin{align}
 \label{HBij}
 \oomoji{H}_{B,i,j}=&[H_{B,i,j}(1),H_{B,i,j}(2),\ldots,H_{B,i,j}(N\cdot N_B^2)]\quad, 
\end{align}
\normalsize
where $H_{B,i,j}(n)$ is the $n$-th element of $\oomoji{H}_{B,i,j}$.

\subsubsection*{Step5.\quad Block Normalization}
Each element of $\oomoji{H}_{B,i,j}$ is normalized as

\footnotesize
\begin{eqnarray}
 \label{normalize}
 \begin{array}{ll}
  H_{B,i,j}(n)=\cfrac{H_{B,i,j}(n)}{\|\oomoji{H}_{B,i,j}\|+\varepsilon}&n=1,2,\ldots,N\cdot N_B^{2}\quad,\label{normalize}\\
 \end{array}
\end{eqnarray}
\normalsize
where $\varepsilon$ is a small constant.
The HOG feature of image $\oomoji{\hat{L}}$ is the vector produced by concanating all normalized HOG blocks.

If HOG features are extracted from EtC images under the conditions of $E^\prime=N_C=E$, $N_B=1$, and $N_O=0$, 
the features are the same as these extracted from unencrypted ones as demonstrated in Sec.\ref{sec:ex}.

\section{Experiment}
\label{sec:ex}

\begin{figure}
 \begin{center}
  \includegraphics[scale=0.45]{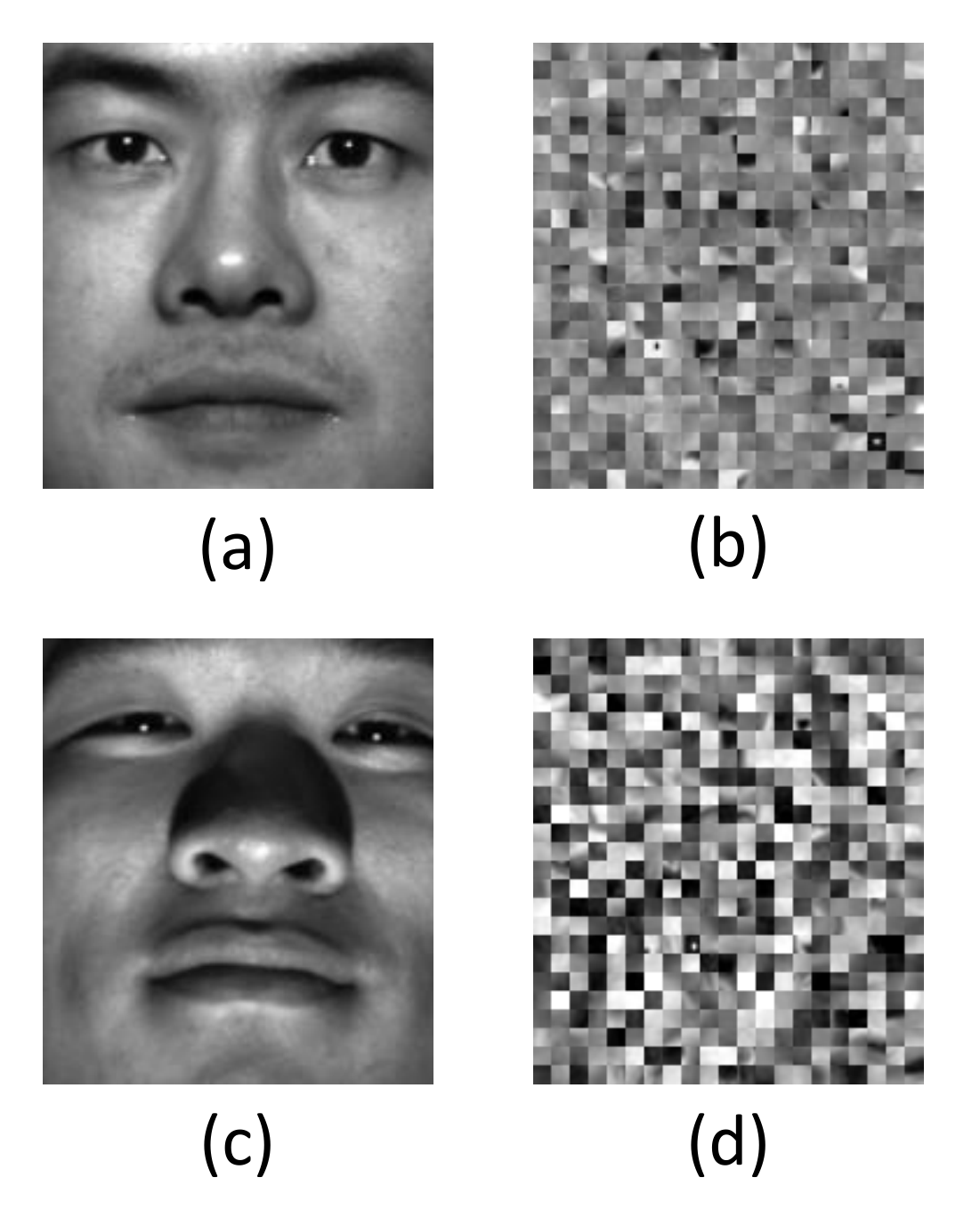}
  \caption{Examples of images from the dataset and the EtC ones.  \label{example yale}}
 \end{center}
\end{figure}

\begin{figure}
 \begin{center}
  \includegraphics[scale=0.6]{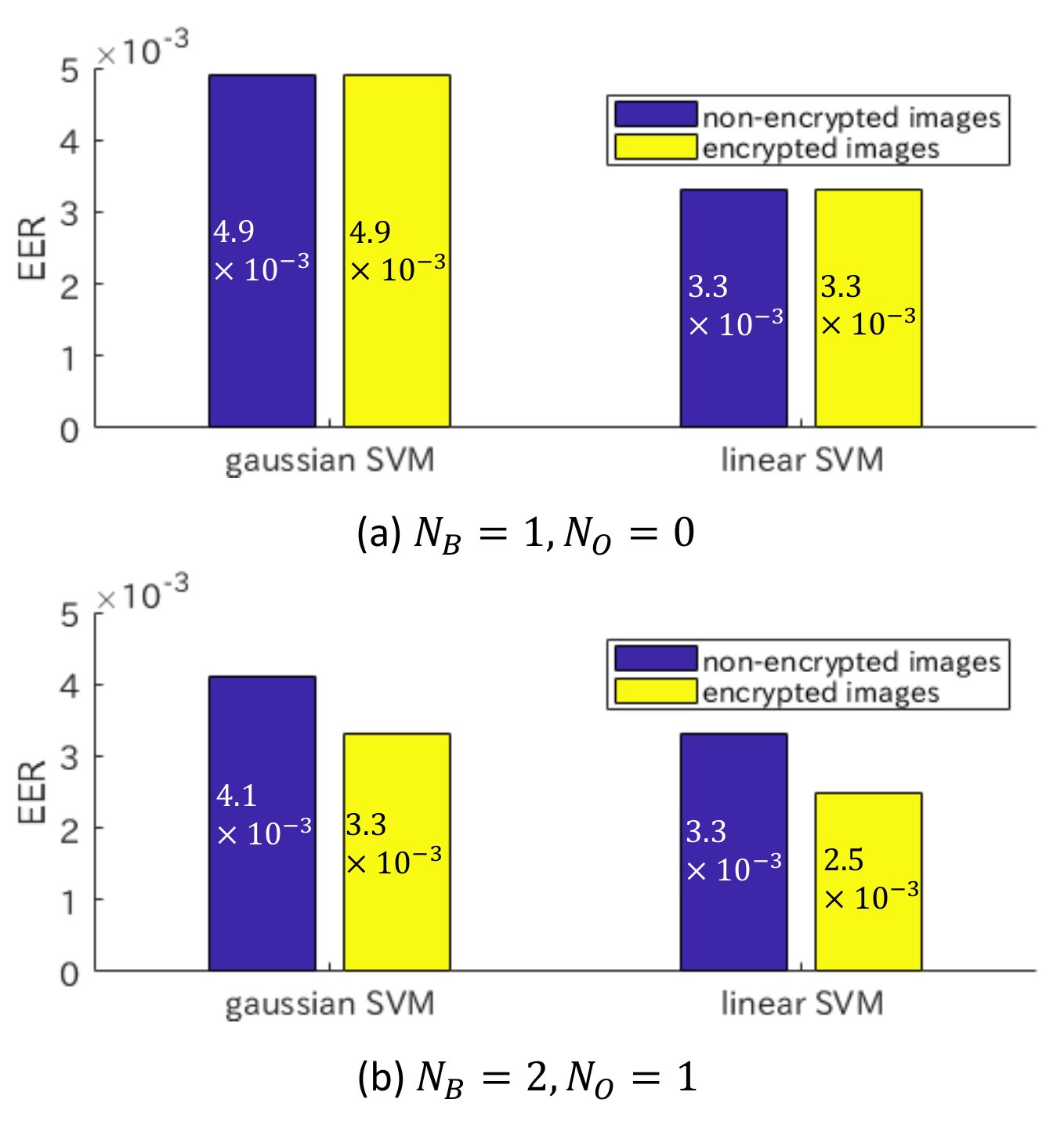}
  \caption{Experiment results ($E^\prime=N_C=E=8,N=10$).  \label{experiment}}
 \end{center}
\end{figure}

A face image classification experiment was carried out under 
the use of linear SVM and gaussian SVM respectively.
We used the Extended Yale Face Database B \cite{Yale}, which consists of 2432 frontal facial images with 168$\times$192 pixels for 38 person.
64 images for each person were divided into half randomly for training data samples and queries.
All the images are encrypted with block size $E=8$.
Figure \ref{example yale} (a) and (c) show example images from the dataset, and Fig.(b) and (d) 
are the EtC images.

Figure \ref{experiment} shows experiment results, 
where HOG feature extraction was carried out under 
two parameter conditions as shown in (a) and (b).
To evaluate the effectiveness of the proposed method, equal error rate (EER), which is the point at which false reject rate (FRR) is equal to false accept rate (FAR), was used.
EER is acquired by changing the threshold of classification score.
A lower EER value means the classifier has a better performance.

Figure \ref{experiment} (a) indicates that the encryption did not give any effect to the  performance of SVM algorithms under condition (a).
Fig.\ref{experiment} (b) indicates  
the performance of SVM algorithms
depends on parameter conditions such as $N_B$ and $N_O$.
Note that the figure also shows the good selection of $N_B$ and $N_O$ can improve 
the classification performance even when the encryption is carried out.

\section{concrusion}
In this paper, we proposed an extraction method of HOG features from images encrypted by a block-based encryption method for privacy-preserving machine learning.
We experimentally showed that the encryption did not give any effect to the performance of machine learning algorithms with HOG features extracted by the proposed method under some conditions exprained in this paper.

\bibliographystyle{IEEEbib}
\bibliography{ref_kitayama20190325}

\end{document}